\documentclass{article}

\PassOptionsToPackage{numbers, compress}{natbib}
\usepackage[preprint]{neurips_2024}

\usepackage[utf8]{inputenc} % allow utf-8 input
\usepackage[T1]{fontenc}    % use 8-bit T1 fonts
\usepackage{hyperref}       % hyperlinks
\usepackage{url}            % simple URL typesetting
\usepackage{booktabs}       % professional-quality tables
\usepackage{amsfonts}       % blackboard math symbols
\usepackage{nicefrac}       % compact symbols for 1/2, etc.
\usepackage{microtype}      % microtypography
\usepackage{graphicx,amsmath,multirow,pifont,caption}
\newcommand{\cmark}{\ding{51}}%
\newcommand{\xmark}{\ding{55}}%

\usepackage{newfloat}
\usepackage{listings}
\usepackage{xurl}
\usepackage{xparse}
\usepackage{array}
\usepackage{stfloats}

\usepackage{helvet}
\usepackage{courier}
\usepackage{colortbl}  
\usepackage[table,xcdraw]{xcolor}
\usepackage[normalem]{ulem}
\usepackage{ragged2e}

\usepackage{color}

\usepackage{rotating}
\usepackage{enumitem}
\usepackage{wrapfig}

\usepackage{extpfeil, extarrows, bm}

\usepackage{algorithm}
\usepackage{algorithmic}

\renewcommand{\paragraph}[1]{\textbf{#1}\ \ }

\renewcommand\footnotemark{}

\title{Robust Zero-Shot Text-to-Speech Synthesis with Reverse Inference Optimization}

\author{%
  Yuchen Hu$^{1,\dag}$\quad Chen Chen$^{1,\dag \thanks{\dag \ Equal Contribution. Listening examples are at: \url{https://yuchen005.github.io/RIO-TTS-demos/}.}}$ \quad \textbf{Siyin Wang}$^{2}$\quad \textbf{Eng Siong Chng}$^1$ \quad \textbf{Chao Zhang}$^{2}$
\\
  $^1$Nanyang Technological University \quad $^2$Tsinghua University\\
  \texttt{\{yuchen005, chen1436\}@e.ntu.edu.sg}\\
}

\begin{document}
\maketitle
\begin{abstract}
In this paper, we propose \textbf{r}everse \textbf{i}nference \textbf{o}ptimization (RIO), a simple and effective method designed to enhance the robustness of autoregressive-model-based zero-shot text-to-speech (TTS) systems using reinforcement learning from human feedback (RLHF).
To assess the quality of speech produced by the TTS system without human annotations, RIO introduces a novel concept termed as \textit{reverse inference} based on the Bayesian principle, which suggests that a high-quality generated speech should be able to be used as a prompt for subsequent generation using the same TTS model. By leveraging reverse inference as the standard to select exemplars used in RLHF from the speech samples generated by the TTS system itself, RIO steers the subsequent optimization towards a direction of enhancing the TTS robustness. 
The RIO framework, comprising sampling, automatic annotating, and learning, obviates the need for a reward model or pairwise preference data, and significantly improves the stability of zero-shot TTS performance by reducing the discrepancies between training and inference conditions. 
Our experimental results verify that RIO can effectively improve both subjective and objective metrics, including mean opinion scores, word error rates, and speaker similarity. Remarkably, RIO can also diminish the incidence of bad outputs to nearly zero percent, rivalling the robustness when using ground-truth speech as the prompt.

%We propose \textbf{r}everse \textbf{i}nference \textbf{o}ptimization (RIO), a simple yet effective approach to improve the robustness of zero-shot text-to-speech (TTS) system with autoregressive codec modeling. RIO is inspired by the recent advancement in reinforcement learning from human feedback, nevertheless, it empowers the TTS model itself to define the ``preference'', instead of relying on human annotators. To achieve it, we propose a new concept of \textit{reverse inference} based on the Bayesian formula, which indicates that beyond synthesizing high-quality speech, a robust TTS model should be able to leverage this synthesized speech as a prompt for further generation. Reverse inference imposes an underlying but high-level requirement on TTS robustness, thus selecting those representative exemplars from self-generated samples to guide subsequent optimization towards the robust direction. This sampling-annotating-learning pipeline eliminates the need for a reward model or pairwise preference data. Yet, it significantly improves the robustness of zero-shot TTS by mitigating the mismatch problem between training and inference. We conduct both subjective and objective experiments to demonstrate the efficacy of RIO in terms of mean opinion score, word error rate, and speaker similarity. Moreover, RIO reduces the ratio of bad cases to nearly 0\%, which even achieves the performance of ground-truth speech.
\end{abstract}

% We present Reverse Inference Optimization (RIO), a simple yet effective approach to improve the robustness of zero-shot text-to-speech (TTS) system with codec language modeling. RIO is inspired by the recent advancement in reinforcement learning from human feedback, nevertheless, it empowers the TTS model itself to define the "preference", instead of relying on human annotators. To achieve it, we propose a new concept of reverse inference based on Bayesian formula, which indicates that beyond synthesizing high-quality speech, a robust TTS model should be able to leverage this synthesized speech as prompt for further generation. Reverse inference imposes a underlying but high-level requirement on TTS robustness, thus selecting those representative exemplars from self-generated samples to guide subsequent optimization towards robust direction. This sampling-annotating-learning pipeline eliminates the need for reward model or pairwise preference data, yet it significantly improve the robustness of zero-shot TTS by mitigating the mismatch problem between training and inference. We conduct both subjective and objective experiments to demonstrate the efficacy of RIO in terms of mean opinion score, word error rate, and speaker similarity. Moreover, RIO reduces the ratio of bad cases to nearly 0%, which even achieves the performance of ground-truth speech.

\section{Introduction}
\label{sec:intro}
%Large language models (LLMs) have emerged as new beacons in natural language processing and generation, especially in zero-shot and in-context learning tasks~\cite{brown2020language,chatgpt,gpt4,touvron2023llama,touvron2023llama2}. Inspired by the success of LLMs, the next token prediction paradigm has been applied to generative tasks in multiple modalities~\cite{zhan2024anygpt,wu2023next,koh2024generating}. For instance, in text-to-speech (TTS) synthesis, increasing works leverage nerual codec modeling~\cite{neuralcodec,soundstream} to discretize speech signals into acoustic tokens, and then tackle the TTS generation with training a decoder-only language model~\cite{audiolm,Audiopalm,speechgpt,wang2023neural}. With extensive text-speech training pairs, the in-context learning capabilities are also emergent in TTS system, which allows the model to perform transcript-conditioned speech continuation task by providing a short speech prompt~\cite{wang2023neural}. Considering the target speaker can be unseen during training, this capability is termed \textit{zero-shot TTS} and has attracted a surge of research interest in speech community~\cite{vallex,wang2023viola,yang2023instructtts,basetts}.
Large language models (LLMs),  capable of zero-shot and in-context learning, have transformed and unified the research and development of natural language processing tasks~\cite{brown2020language,chatgpt,gpt4,touvron2023llama,touvron2023llama2}. 
Inspired by the success of text LLMs, the next-token prediction paradigm has been generalised to tasks of other modalities~\cite{zhan2024anygpt,wu2023next,koh2024generating}, in particular, text-to-speech (TTS) synthesis implemented as an auto-regressive language model based on the neural codec tokens~\cite{neuralcodec,soundstream} extracted by discretizing the speech signals~\cite{audiolm,Audiopalm,speechgpt,wang2023neural}. 
With extensive text-speech training pairs, the in-context learning capabilities are also emergent in such TTS systems, enabling the model to perform transcript-conditioned speech continuation tasks by providing a short speech prompt~\cite{wang2023neural}. Since the speaker in the speech prompt can be unseen during training, this capability is often termed \textit{zero-shot TTS} and has attracted a surge of research interest in the speech community~\cite{vallex,wang2023viola,yang2023instructtts,basetts}.

% Poor robustness ==> Train-inference mismatch
%Despite the remarkable progress in zero-shot TTS, the codec language model suffers from sub-optimal robustness in auto-regressive generation~\cite{yao2012adaptation,huang2021context}, resulting in unstable speech synthesis, e.g., abrupt truncation and weird prosody~\cite{xin2024ralle}. We attribute this issue to the defective training schedule of codec language model, where the teacher-forcing strategy causes the model's next token prediction to depend on the ground truth sequence~\cite{ji2023survey}. However, during zero-shot codec TTS inference, the model must perform autoregressive decoding based on its own predicted sequence, leading to exposure bias and error accumulation~\cite{bengio2015scheduled,pang2020text,gu2023minillm}.

Despite the significant advancements, the zero-shot TTS often has sub-optimal robustness that results in unstable speech synthesis performance~\cite{yao2012adaptation,huang2021context}, with an unignorable rate of abrupt truncation, repetition, and unnatural prosody~\cite{xin2024ralle}. 
We attribute this problem to the incomplete training of the codec-based auto-regressive language models, which suffer from the issue of \textit{exposure bias} well-known in speech recognition and other sequence modelling problems~\cite{bengio2015scheduled,pang2020text,gu2023minillm}. Specifically, the codec language model is only trained with the teacher-forcing strategy that predicts the next token by depending on the golden history sequence without considering the errors accumulated from the history in auto-regressive decoding at test time~\cite{ji2023survey}.

Recently, there has been a growing interest in integrating human evaluation into TTS optimization through reinforcement learning from human feedback (RLHF)~\cite{rafailov2024direct}, which effectively enhances the zero-shot capacity of pre-trained TTS models~\cite{speechalign,chen2024uno}.
RLHF typically follows a sampling-annotating-learning pipeline, in which human evaluation is applied to model-generated outputs to ensure they align with subjective human preferences. We suggest that this pipeline offers a viable strategy to address the discrepancy in history sequence between TTS training and test, as it exposes the model with the samples with self-generated history during training~\cite{ethayarajh2024kto}. 
To effectively handle such exposure bias issue in zero-shot TTS using RLHF, it is crucial to find a suitable ``preference'' function that can efficiently determine whether selecting a specific speech sample as a positive exemplar can lead to improved TTS robustness or not in RLHF framework.

\begin{figure*}[t]
  \begin{center}
  \includegraphics[scale=1.3]{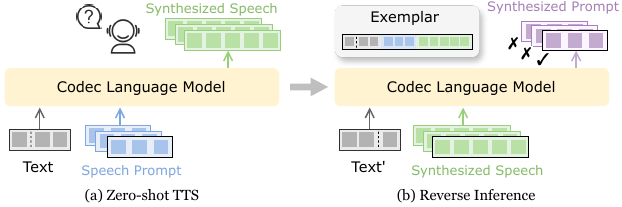}
  \end{center}
  \caption{The overview of RIO. 
    \textbf{(a) Zero-shot TTS:} Codec language model generates the synthesized speech conditioned on a 3-second speech prompt and text (including both text prompt and transcription), where the synthesized speech could be high-quality but \textit{not necessarily perceptually consistent with its speech prompt}.
  \textbf{(b) Reverse inference:} The synthesized speech is sent back to the TTS model to predict the original prompt speech, whose quality can reflect the production-perception consistency (PPC) of previously synthesized speech.
  We then set those PPC samples as positive exemplars in RLHF to optimize the TTS model towards better robustness.
%  \textbf{(a) Zero-shot TTS:} Codec language model generates the synthesized speech conditioned on a 3-second speech prompt and text (including both text prompt and transcription), where the synthesized speech could be high-quality but \textit{not necessarily robust}.
%  \textbf{(b) Reverse inference:} The synthesized speech is sent back to the TTS model to predict the prompt speech, whose quality can in turn reflect the robustness of previously synthesized speech.
%  We then set those ``robust'' samples as positive exemplars in RLHF to optimize the TTS model with stronger robustness.
  }
  \vspace{-0.2cm}
  \label{f1}
\end{figure*}

In this paper, we introduce \textbf{r}everse \textbf{i}nference \textbf{o}ptimization (RIO), an RLHF-related method tailored to improve the robustness of zero-shot TTS, which alleviates the train-inference mismatch through a sampling-annotating-learning pipeline based on a novel self ``preference'' function. 
Our proposed self ``preference'' function is built on \textit{reverse inference} defined based on the Bayesian formula, which relies on the assumption that a satisfactory speech sample generated by a robust zero-shot TTS system should serve as a good speech prompt to \textit{reversely} generate the original speech prompt using the same TTS system. %This assumption imposes an underlying requirement of \textit{perceptual consistency} that requires the TTS-produced speech samples to be perceived consistently as the human-produced speech samples by the TTS system as a speech prompt. 
This assumption imposes an underlying requirement of \textit{production-perception consistency} (PPC) that enforces the TTS-produced speech samples to be consistent with the human-produced speech samples when perceived by the TTS system as the speech prompt, which we humans can not notice. To this end, the self ``preference'' is defined as selecting the speech samples satisfying both forward and reverse TTS inference as the positive exemplars for the subsequent learning process. 
Besides the target TTS model, a pre-trained mean opinion score (MOS) estimator is used to assess the quality of a speech sample, which can thereby avoid the intensive human annotation in traditional RLHF. 
In addition, RIO eliminates the requirement of either reward model or pairwise preference data by directly maximizing the utility of speech generations to improve PPC in both the forward and reverse inference. 
Experimental results show that RIO considerably enhances the robustness of zero-shot TTS, and the quality of the synthesized speech is improved consistently in terms of MOS, word error rate (WER), speaker similarity (SIM), and the ratio of bad cases.

Our main contributions are summarized as follows:
\begin{itemize}
    \item By investigating the impact of synthesized speech prompts on zero-shot TTS, we propose and verify an assumption based on Bayesian reverse inference that PPC samples are correlated well with TTS robustness, providing new insights for understanding the difference between human and machine speech perception.      
    \item RIO, a sampling-annotating-learning pipeline is proposed to improve the TTS robustness by mitigating training-inference mismatch. The novelty of RIO lies in the exemplar selection strategy that utilizes the self ``preference'' provided by the TTS model to determine whether a speech sample is suitable for subsequent RLHF optimization. 
    %, instead of relying on human evaluation.  
    \item Intensive experiments are conducted to demonstrate the efficacy of RIO in both subjective and objective metrics. 
    In particular, RIO reduces the ratio of bad cases to nearly 0\%, which effectively resolves the robustness issue of codec-based auto-regressive zero-shot TTS. 
    
    %demonstrates its superior robustness over baselines.
\end{itemize}

%Our main contributions are summarized as follows:
%\begin{itemize}
%    \item We investigate the impact of speech prompts on zero-shot TTS and propose a correlation between Bayesian reverse inference and robustness, providing new insights for analyzing sequential TTS decoding in codec language models. 
%    \item RIO, a sampling-annotating-learning pipeline is proposed to improve the TTS robustness by mitigating training-inference mismatch. RIO is novel for the exemplar selection strategy that utilizes the TTS model itself to determine which speech samples are suitable for subsequent optimization, instead of relying on human evaluation.  
%    \item We conduct intensive experiments to demonstrate the efficacy of RIO in terms of multiple objective and subjective metrics.    Moreover, RIO is shown able to reduce the ratio of bad cases to nearly 0\%, which demonstrates its superior robustness over baselines.
%\end{itemize}

% Humans are not precise evaluators of speech—our auditory system often struggles to perceive subtle differences in speech and provide reliable judgments.  

\section{Related Work}
\textbf{Neural codec language modeling} formulates TTS generation as a next-token prediction task and has gained increasing popularity in recent years~\cite{audiolm,Audiopalm,speechgpt}. Under this setup, the speech signal is firstly tokenized into sequences of discrete units based on vector quantization~\cite{neuralcodec,soundstream,zhang2023speechtokenizer,huang2023repcodec}, and then a decoder-only language model is trained based on these acoustic tokens~\cite{mohamed2022self,TortoiseTTS}. This approach has been demonstrated promising scalability to large data and model sizes, resulting in high-quality synthesized speech~\cite{wang2023neural,Spear-TTS}, emergent zero-shot capacity on unseen speakers~\cite{voicecraft,basetts,xin2024ralle}, style control~\cite{lyth2024natural,ji2024textrolspeech,liu2023promptstyle,yang2023instructtts}, and cross-lingual TTS~\cite{vallex,wang2023viola}.

\textbf{RLHF-based optimization} is widely utilized for LLM alignment, where a reward model is trained with human-annotated data to calibrate the generative content~\cite{christiano2017deep,bai2022training, achiam2023gpt,dai2023safe}. Recent advancements focus on closed-form losses that directly operate on preference data, such as direct preference optimization (DPO)~\cite{rafailov2024direct} and its extensions~\cite{amini2024direct,zeng2024token,liu2024enhancing}. Moreover, the idea of ``self-rewarding'' is also proposed to infer preference data based on the model itself~\cite{chen2024self,yuan2024self}. 
In the realm of TTS optimization, SpeechAlign~\cite{speechalign} presents the first method based on DPO that views ground truth as preferred samples while model generations as dispreferred samples. 
UNO~\cite{chen2024uno} eliminates the dependence on such pairwise preference data and considers the annotation uncertainty due to subjective variability.

\section{Methodology}
\subsection{Problem Formulation of Zero-shot TTS}
Given speech prompt $\mathbf{X}$ and its paired text prompt $\textbf{T}_\text{X}$, zero-shot TTS aims to synthesize a target speech $\mathbf{Y}$ based on target text $\textbf{T}_\text{Y}$ that clones the voice of the speaker in $\mathbf{X}$. With neural codec modelling, both $\mathbf{X}$ and $\mathbf{Y}$ are represented as a sequence of discrete acoustic tokens, and the auto-regressive inference process can be formulated as speech continuation that a trained codec language model predicts most possible speech sequence $\hat{\mathbf{Y}}$:
% \begin{equation}
%     s_l = \pi_{\theta_1} (t, p, s_{<l}), \ l\in \{1,2,\ldots,L\},
% \label{eq-tts-infer}
% \end{equation}
\begin{equation}
    \hat{\mathbf{Y}} = \arg\max\nolimits_\textbf{Y}P (\mathbf{Y} | \textbf{T}_\text{Y}, \textbf{T}_\text{X}, \mathbf{X}),
\label{eq-tts-infer}
\end{equation}
and $\hat{\mathbf{Y}}$ typically consists of tokens of several codebooks from different residual vector quantizers (RVQs)~\cite{neuralcodec}. 
The tokens from the first quantizer are predicted in an auto-regressive manner, and then refined in a non-autoregressive manner to reconstruct the time-domain waveform~\cite{wang2023neural}.  

\subsection{Reverse Inference}
\label{ssec:reverse_infer}
%Given speech $\hat{\mathbf{Y}}$ synthesized by the TTS model, a natural question is whether $\hat{\mathbf{Y}}$ can be further utilized as a prompt for TTS inference. We hereby define the reverse inference process that takes $\hat{\mathbf{Y}}$ as input to predict the original speech prompt $\mathbf{X}$:
For speech $\hat{\mathbf{Y}}$ generated by a text-to-speech (TTS) model, it is pertinent to explore whether $\hat{\mathbf{Y}}$ can serve as a prompt for further TTS inference. This leads us to investigate a ``reverse'' inference process in which $\hat{\mathbf{Y}}$ is used as input to predict the original speech prompt $\mathbf{X}$.
Hereby, we propose a reverse inference that takes $\hat{\mathbf{Y}}$ and $\textbf{T}_\text{X}$ as the speech prompt and target text to synthesize a speech sample $\hat{\mathbf{X}}$: 
\begin{equation}
    \hat{\mathbf{X}} = \arg\max\nolimits_{\textbf{X}} P (\mathbf{X} | \textbf{T}_\text{X}, \textbf{T}_\text{Y}, \hat{\mathbf{Y}}),
\label{eq-tts-reverse}
\end{equation}
where the original target text $\textbf{T}_\text{Y}$ and the transcription of original speech prompt $\textbf{T}_\text{X}$ are swapped accordingly. Eq.~\eqref{eq-tts-infer} and Eq.~\eqref{eq-tts-reverse} can be connected using Bayes’ theorem as:
\begin{align}
P(\mathbf{Y} | \textbf{T}_\text{Y}, \textbf{T}_\text{X}, \mathbf{X}) &= \frac{P(\mathbf{X}| \textbf{T}_\text{X}, \textbf{T}_\text{Y},\mathbf{Y}) \, P(\mathbf{Y} |\textbf{T}_\text{Y}, \textbf{T}_\text{X})}{P(\mathbf{X}| \textbf{T}_\text{X}, \textbf{T}_\text{Y})} \label{eq-bayes1} \\
&= \frac{P(\mathbf{Y} |\textbf{T}_\text{Y})}{P(\mathbf{X}| \textbf{T}_\text{X})}P(\mathbf{X}| \textbf{T}_\text{X}, \textbf{T}_\text{Y}, \mathbf{Y})
\label{eq-bayes2}
\end{align}
where $(\textbf{T}_\text{X}, \mathbf{X})$ and $(\textbf{T}_\text{Y}, \mathbf{Y})$ are independent text-speech pairs, and  
$P(\mathbf{X}|\textbf{T}_\text{X})$ and $P(\mathbf{Y}|\textbf{T}_\text{Y})$ are the priors of these pairs that can be viewed as constant. 
%Since $\mathbf{Y}$ is not available during inference, the likelihood $P(\mathbf{X}| T_2, T_1, \mathbf{Y})$ is approximated by $P(\mathbf{X}| T_2, T_1, \hat{\mathbf{Y}})$. As a result, the reverse inference process exhibits a direct correlation with zero-shot inference. \par
From Eq.~\eqref{eq-bayes2}, training the TTS model to maximize $P(\mathbf{Y}|\textbf{T}_\text{Y},\textbf{T}_\text{X}, \mathbf{X})$ should maximize $P(\mathbf{X}| \textbf{T}_\text{X},\textbf{T}_\text{Y}, \mathbf{Y})$ simultaneously, which should naturally enable $P(\mathbf{X}| \textbf{T}_\text{X},\textbf{T}_\text{Y}, \hat{\mathbf{Y}})$ if $\hat{\mathbf{Y}}$ is a sufficiently good approximation to $\mathbf{Y}$. 

%Since $\mathbf{Y}$ is not available during inference, the likelihood $P(\mathbf{X}|\textbf{T}_\text{X},\textbf{T}_\text{Y}, \mathbf{Y})$ is approximated by $P(\mathbf{X}| \textbf{T}_\text{X},\textbf{T}_\text{Y}, \hat{\mathbf{Y}})$. As a result, the reverse inference process exhibits a direct correlation with zero-shot inference.

%The reverse inference capability indicates that the TTS model can understand its self-generated content and perform secondary prediction based on it. Not only should the synthesized speech be satisfactory, it should also be a prefix suitable for future stable generation. Intuitively, this capability imposes a higher level of robustness requirement on the model. We design a quick experimental validation for the following assumption: Given two TTS models $\theta_1$ and $\theta_2$, if their zero-shot performances on a small set $\mathcal{D}_{\text{val}}$ are comparable, the model with better robustness will exhibit superior reverse inference capability. In practice, we employ a pre-retained \textit{MOSNet} as discriminator in terms of estimated MOS. $\theta_1$ and $\theta_2$ are respectively 830M and 330M versions of VoiceCraft model with remarkable zero-shot capacity. $\mathcal{D}_{\text{val}}$ is sampled from LibriSpeech~\cite{panayotov2015librispeech}, including 100 data points that $\theta_1$ and $\theta_2$ achieve MOS from 3.8 to 4.8. The results of reverse inference are shown in Figure~\ref{f2} and Table~\ref{table:mos_330_830}, where the threshold of bad case is set as 3.0.

\begin{figure}
  % \centering
    \begin{minipage}[ht]{0.57\textwidth}
    \centering
    \includegraphics[scale=0.22]{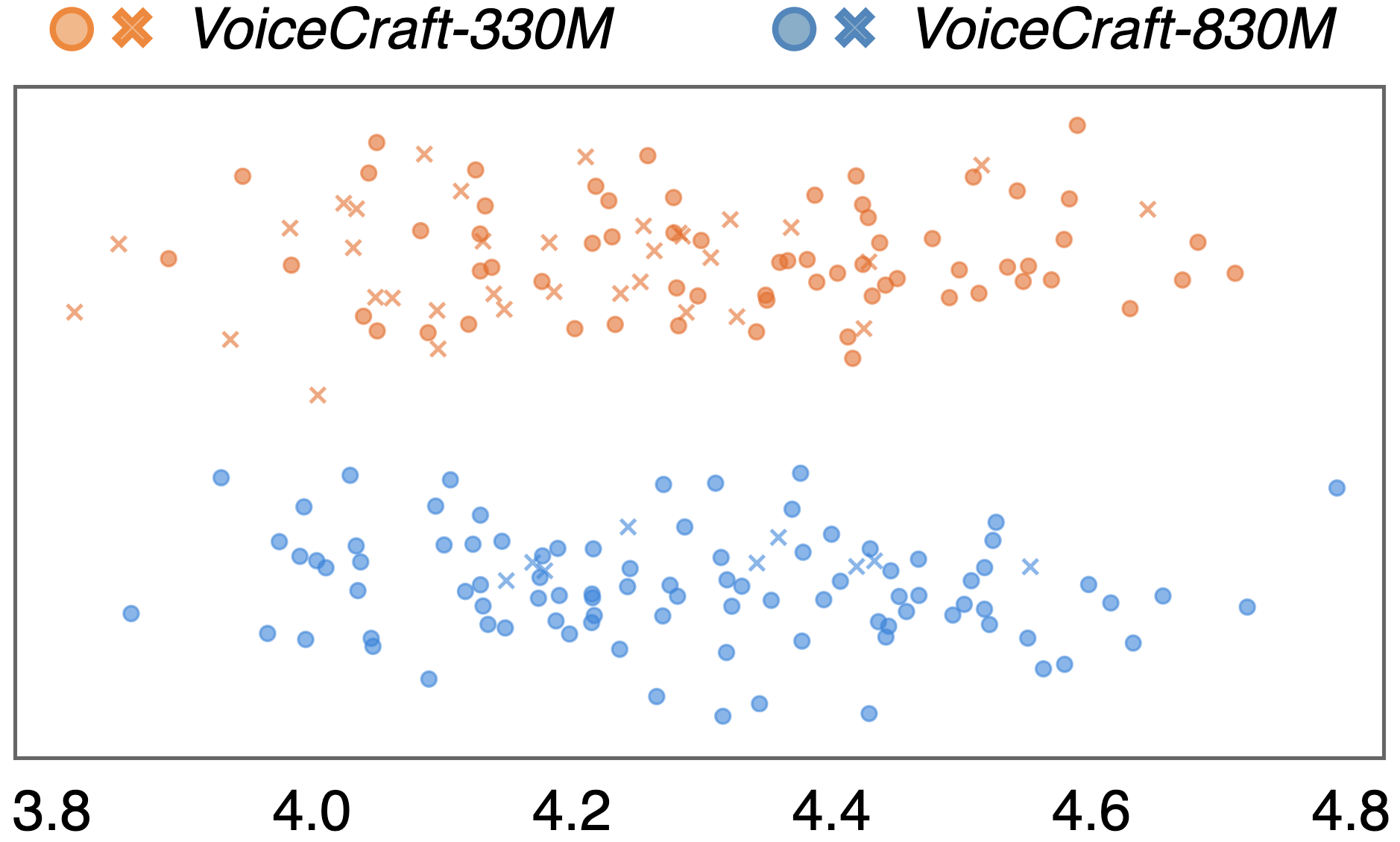}
    \caption{MOS distributions of $\hat{\mathbf{Y}}$ synthesized by different models using zero-shot generations with MOS > 3.8. The circle/cross denotes good/bad reverse inference results.}
    \label{f2}
    \end{minipage}
  \hfill
    \begin{minipage}[ht]{0.4\textwidth}
    \centering
    \captionof{table}{Averaged MOS of good zero-shot generations $\hat{\mathbf{Y}}$ that have different reverse inference results $\hat{\mathbf{X}}$. 
    ``\cmark\xmark'' means good zero-shot but bad reverse inference results, and ``\cmark\cmark'' means both results are good (with MOS > 3).
    \textit{VoiceCraft} models of different sizes (330M and 830M) are used for statistics.
    }
    \label{table:mos_330_830}
    \scalebox{1.0}{
    \begin{tabular}{c|c|c} \toprule 
    Model & \cmark\xmark & \cmark\cmark  \\ 
    \midrule
    \text{VoiceCraft-330M} & $3.79$   &   $3.90$ \\ \midrule
    \text{VoiceCraft-830M}   & $4.24$   &   $4.32$ \\
    \bottomrule
    \end{tabular}}
  \end{minipage}
\end{figure}

\textbf{Empirical observation} shows that when applying pre-trained zero-shot TTS models to reverse inference, the $\hat{\mathbf{X}}$ generated based on Eq.~\eqref{eq-tts-reverse} sometimes has very low quality even if the corresponding $\hat{\mathbf{Y}}$ has high quality, in both subjective and objective metrics. 
Specifically, two TTS models, VoiceCraft with 330 million (M) and 830M parameters referred to as $\bm{\theta}_1$ and $\bm{\theta}_2$, with similar zero-shot performances on a small validation dataset $\mathcal{D}_{\text{val}}$ are examined. $\mathcal{D}_{\text{val}}$ consists of 1000 data samples selected from the LibriSpeech dataset~\cite{panayotov2015librispeech}, whose corresponding synthetic speech samples produced by both $\bm{\theta}_1$ and $\bm{\theta}_2$ satisfy MOS > 3 (estimated by the MOSNet). 
The results of the samples $\hat{\mathbf{X}}$ obtained by reverse inference are shown in Figure~\ref{f2} and Table~\ref{table:mos_330_830}. 

Although both $\bm{\theta}_1$ and $\bm{\theta}_2$ generate good $\hat{\mathbf{Y}}$ with high MOS on $\mathcal{D}_{\text{val}}$, $\bm{\theta}_2$ generates much fewer bad $\hat{\mathbf{X}}$ (with MOS $\leqslant$ 3) compared to $\bm{\theta}_1$ ($9\%$ {vs.} $35\%$). This phenomenon reveals the underlying difference between human-produced speech samples and high-quality TTS-produced speech samples, which may be attributed to subtle changes that human listeners cannot perceive. %To resolve this issue, we propose to enhance TTS robustness 
Therefore, to further reduce the discrepancy between the speech samples produced by humans and TTS, we propose to further optimize the TTS towards generating high-quality $\hat{\mathbf{X}}$, which requires the TTS speech production to be consistent with its own perception of the speech prompt. The next section will present how to achieve such consistency using RLHF with the PPC samples, which also reduces exposure bias.

%\textbf{Empirical Observation.} Though $\theta_1$ and $\theta_2$ achieve comparable MOS on $\mathcal{D}_{\text{val}}$, the robust model $\theta_1$ generate much less bad cases than $\theta_2$ ($9\%$ \textit{vs.} $35\%$). This phenomenon potentially reveals the fact that humans are not precise speech evaluators, as our auditory system can not perceive subtle differences in speech generations. However, the reverse inference of TTS models can provide underlying differentiation for samples with similar MOS in $\mathcal{D}_{\text{val}}$. This is crucial for selecting samples for subsequent training -- they serve as exemplars guiding the optimization direction of the TTS model.

\subsection{Optimization without Pairwise Preference Data}
With $K$ times of zero-shot TTS sampling, $I$ PPC speech samples with both good inference and reverse inference results are selected as exemplars. Conversely, $J$ bad cases are also selected as negative examples to inhibit generating such undesirable samples. In this process, {MOSNet} is employed as the discriminator according to a threshold (details are in Sec.~\ref{dataset}), hence $K$ can be large to provide representative samples. The bad cases are independent of the exemplars since they are obtained with different inputs, which are stored in a positive pool $\mathcal{P}_{\text{pos}}$ and a negative pool $\mathcal{P}_{\text{neg}}$ separately by:      
%With $K$ times of zero-shot TTS sampling, $I$ PPC speech samples with both good inference and reverse inference results are selected as exemplars. Conversely, $J$ bad cases are also selected as negative examples to suppress such undesirable generation. In this process, the \textit{MOSNet} is employed as the discriminator according to a threshold (details are in \S\ref{dataset}), hence $K$ can be large to provide representative samples. Furthermore, these bad cases are independent of the exemplar since they consume different inputs, and we store them using a positive pool $\mathcal{P}_{\text{pos}}$ and a negative pool $\mathcal{P}_{\text{neg}}$:       
\begin{align}
    \mathcal{P}_{\text{pos}} &= \{(\mathbf{X}_i, \hat{\mathbf{Y}}_i, \textbf{T}_{\text{X},i}, \textbf{T}_{\text{Y},i}) \ | \ \hat{\mathbf{Y}}_i \sim \pi_{\text{ref}} (\mathbf{X}_i, \textbf{T}_{\text{X},i}, \textbf{T}_{\text{Y},i}),i = 1,2, \ldots, I\} \\
    \mathcal{P}_{\text{neg}} &= \{(\mathbf{X}_j, \hat{\mathbf{Y}}_j, \textbf{T}_{\text{X},j}, \textbf{T}_{\text{Y},j} \ | \ \hat{\mathbf{Y}}_j \sim \pi_{\text{ref}} (\mathbf{X}_j, \textbf{T}_{\text{X},j}, \textbf{T}_{\text{Y},j}),j = 1,2, \ldots, J\},
\end{align}
where $\pi_{\text{ref}}$ denotes a frozen reference model that prevents the optimized model $\pi_{\theta}$ from making radical update. Since the samples in $\mathcal{P}_{\text{pos}}$ and $\mathcal{P}_{\text{pos}}$ are \textit{not} pairwise preference data, a ``reference point'' is used following prior works~\cite{ethayarajh2024kto}, which is estimated by a KL divergence item $\mathcal{Z}_\text{kl}$. That is,
\begin{align}
\label{reference-point}
    \mathcal{Z}_{\text{kl}} = \mathbb{E}_{(\mathbf{X}, \hat{\mathbf{Y}}, \textbf{T}_{\text{X}}, \textbf{T}_{\text{Y}})\sim \mathcal{P}_{\text{pos}}\cup \mathcal{P}_{\text{neg}}} [ \text{KL}(\pi_{\theta}(\hat{\mathbf{Y}}|\mathbf{X}, \textbf{T}_{\text{X}}, \textbf{T}_{\text{Y}}) \Vert \pi_{\text{ref}}(\hat{\mathbf{Y}}|\mathbf{X}, \textbf{T}_{\text{X}}, \textbf{T}_{\text{Y}}) )],
\end{align}
where $(\mathbf{X}, \hat{\mathbf{Y}}, \textbf{T}_{\text{X}}, \textbf{T}_{\text{Y}})$ samples from each batch during training. Based on $\mathcal{Z}_\text{kl}$, the final optimization loss for the TTS system is written as:   
\begin{align}  
\mathcal{L}_{\text{tts}} (\pi_{\theta}, \pi_{\text{ref}})&= \mathbb{E}_{(\mathbf{X}, \hat{\mathbf{Y}}, \textbf{T}_{\text{X}}, \textbf{T}_{\text{Y}})\sim \mathcal{P}_{\text{pos}}\cup \mathcal{P}_{\text{neg}}} (1-\mathcal{V}_{\text{tts}}(\mathbf{X}, \hat{\mathbf{Y}}, \textbf{T}_{\text{X}}, \textbf{T}_{\text{Y}})) \label{eq-loss}\\
\mathcal{V}_{\text{tts}}(\mathbf{X}, \hat{\mathbf{Y}}, \textbf{T}_{\text{X}}, \textbf{T}_{\text{Y}}) & = 
\begin{cases}
\sigma(\mathcal{R}(\mathbf{X}, \hat{\mathbf{Y}}, \textbf{T}_{\text{X}}, \textbf{T}_{\text{Y}}) - \mathcal{Z}_{\text{kl}}), & \text{if } (\mathbf{X}, \hat{\mathbf{Y}}, \textbf{T}_{\text{X}}, \textbf{T}_{\text{Y}}) \sim \mathcal{P}_{\text{pos}} \\
\sigma(\mathcal{Z}_{\text{kl}} - \mathcal{R}(\mathbf{X}, \hat{\mathbf{Y}}, \textbf{T}_{\text{X}}, \textbf{T}_{\text{Y}})), & \text{if } (\mathbf{X}, \hat{\mathbf{Y}}, \textbf{T}_{\text{X}}, \textbf{T}_{\text{Y}}) \sim \mathcal{P}_{\text{neg}} 
\end{cases} \label{eq-value} \\
\mathcal{R}(\mathbf{X}, \hat{\mathbf{Y}}, \textbf{T}_{\text{X}}, \textbf{T}_{\text{Y}}) &= \beta \cdot \log \frac{\pi_\theta(\hat{\mathbf{Y}} | \mathbf{X}, \textbf{T}_{\text{X}}, \textbf{T}_{\text{Y}})}{\pi_{\text{ref}}(\hat{\mathbf{Y}} | \mathbf{X}, \textbf{T}_{\text{X}}, \textbf{T}_{\text{Y}})}. \label{eq-reward}
\end{align}
In Eq.~\eqref{eq-reward}, $\mathcal{R}(\mathbf{X}, \hat{\mathbf{Y}}, \textbf{T}_{\text{X}}, \textbf{T}_{\text{Y}})$ is the implicit reward modeling proposed in DPO~\cite{rafailov2024direct}, where $\beta$ serves as a factor controlling update step from $\pi_{\text{ref}}$. Eq.~\eqref{eq-value} reflects the core idea of optimization: Given a desirable sample from $\mathcal{P}_{\text{pos}}$, the corresponding implicit reward is maximized to boost the probability of $\pi_{\theta}$, conversely, an undesirable sample is suppressed to avoid such kind of inference. In this process, $\mathcal{Z}_\text{kl}$ is not involved in the backpropagation but it stabilizes the training process. $\sigma(\cdot)$ is the logistic function to keep $\mathcal{V}_{\text{tts}}$ less than 1, and minimizing the $\mathcal{L}_{\text{tts}}$ in Eq.~\eqref{eq-loss} is equal to maximize the $\mathcal{V}_{\text{tts}}$. \par

Here the advantages of RIO in summary:
(1) RIO only requires a desirable/undesirable label for each sample, thus supporting flexible annotating and eliminating the need for pairwise preference data. Based on the sample input $(\mathbf{X}, \textbf{T}_\text{X}, \textbf{T}_\text{Y})$, it is challenging for the same TTS model to simultaneously generate both a PPC exemplar and a bad sample to constitute the pairwise data required by DPO.
(2) Since no human evaluation is required in this process, the sampling frequency $K$ can be increased significantly to ensure that the $\mathcal{P}_{\text{pos}}$ and $\mathcal{P}_{\text{neg}}$ contain a sufficient number of high/bad-quality examples.

\section{Experimental Setup}

\textbf{Dataset.}
\label{dataset}
The data used in our experiments consists of three parts: supervised training for the backbone TTS model, optimization with RIO, and evaluation. 
There are no overlapping speakers among them. 
(1) Our backbone model VoiceCraft~\cite{voicecraft} is trained on GigaSpeech dataset~\cite{chen2021gigaspeech} that contains 9k hours of audiobooks, podcasts, and YouTube videos at sampling rate of 16kHz. 
(2) RIO is built on LibriTTS~\cite{zen2019libritts} dataset that has no overlapping with Gigaspeech.
Following previous work~\cite{chen2024uno}, we sample a pool of speech prompts with audio files around 3 seconds (commonly used in zero-shot TTS studies), and then perform zero-shot TTS generation based on another pool of target transcripts containing over 6 tokens.
As a result, we obtained 2,000 training samples, and then we selected 200 positive samples and 200 negative samples according to RIO policy for RLHF optimization.
Specifically, we first calculate the average MOS of zero-shot inference and reverse inference results, where the samples with top average MOS are set as positive samples and those with bottom average MOS are set as negative samples.
Then, we use the WER metric to further refine the quality of exemplars, where only the positive samples with WER lower than 10\% and the negative samples with WER higher than 10\% are maintained.
(3) For evaluation, we use a subset from LibriSpeech test-clean~\cite{panayotov2015librispeech} with the audio lengths ranging from 5 to 16 seconds (longer samples which are harder for autoregressive models than in previous work~\cite{chen2024uno} and thus better evaluate the robustness of zero-shot TTS).
Same as the training process, for each test sample we randomly select a 3-second prompt speech with the same speaker identity from another non-overlapping pool.

\textbf{Models.} 
VoiceCraft\footnote{\url{https://huggingface.co/pyp1/VoiceCraft/tree/main}}~\cite{voicecraft} is used as the backbone model due to its demonstrated superior zero-shot TTS capability, where both base (330M) and large (830M) pre-trained models are used as our starting points.
VoiceCraft adopts the popular codec language model scheme~\cite{wang2023neural}, where the pre-trained Encodec contains 4 RVQ codebooks with 2,048 code entries.
RIO finetunes all parameters of pre-trained VoiceCraft with learning rate set to 1e-5 and batch size set to 2. 
AdamW~\cite{loshchilov2018decoupled} is used as the optimizer and only trains 1 epoch, which takes 10 minutes on a single NVIDIA-A100-40GB GPU.

\textbf{Objective Evaluation.} 
Following prior studies, WER and SIM are used to evaluate speech intelligibility and speaker information respectively.
WER is calculated using pre-trained Whisper-medium.en\footnote{\url{https://huggingface.co/openai/whisper-medium.en}} speech recognition model, and SIM is calculated using pre-trained WavLM-TDCNN\footnote{\url{https://huggingface.co/microsoft/wavlm-base-plus-sv}} speaker recognition model.
For evaluation of speech naturalness, we use the open-sourced {MOSNet}\footnote{\url{https://github.com/nii-yamagishilab/mos-finetune-ssl}}~\cite{cooper2022generalization} to estimate an objective score of MOS for reference, which has been demonstrated with good generalization capability to out-of-domain data.
For subjective evaluation, we also invite human listeners to evaluate our proposed approach and comparison baselines.

\textbf{Human Evaluation.} 
We randomly sample 20 listening examples from 5-8 seconds, 8-12 seconds, and 12-15 seconds, respectively, to cover different lengths of samples.
Then, six human listeners are invited to assess these 60 synthesized speech samples using MOS.
Listeners are tasked with rating the naturalness of each audio sample on a 5-point Likert scale, ranging from 1 (very unnatural) to 5 (completely natural).
Furthermore, these speech samples are randomly assigned to listeners for side-by-side A/B testing (10 samples per listener). 
After listening to two generated speech samples from different models but same prompt, listeners are asked to choose one that sounds more natural, or if they are too close to distinguish, indicating a tie. 

\textbf{Baselines.} Apart from the backbone {VoiceCraft} model by typical supervised learning, we also reproduce the following optimization approaches based on {VoiceCraft} for comprehensive comparison:
\begin{itemize}
    \item \textit{RIO-DPO}: We implement the popular DPO algorithm~\cite{rafailov2024direct} from RLHF community in our proposed RIO framework.
    Different from RIO optimization, this baseline requires paired data where positive and negative samples are generated using the same TTS input.
    To this end, we repeat 5 times of TTS generation for each transcription and only select those with a large gap of average MOS (\textit{i.e.}, larger than 2) between the best and worst generations.
    \item \textit{RIO-ODPO}: Recent work~\cite{amini2024direct} presents ODPO, an enhanced version of the DPO algorithm with a considering offset that shows good effectiveness. Following them, we add the gap between the average MOS of paired positive and negative samples as the ``offset'' on top of the \textit{RIO-DPO} baseline to achieve ODPO optimization.
    \item \textit{UNO-null}: Recent work~\cite{chen2024uno} proposes an uncertainty-ware optimization method within the RLHF framework to improve the subjective performance of zero-shot TTS, which achieves superior MOS results over all baselines. For a fair comparison, we discard the extra uncertainty information and reproduce {UNO-null} as a non-reverse-inference version of RIO.
    \item \textit{Ground-Truth}: Since ground-truth speech samples of the test set are available, we evaluate their corresponding metrics as the upper bound for reference. 
\end{itemize}

\section{Results and Analysis}

\begin{table}[t]
\caption{Main results on word error rate (WER), speaker similarity (SIM), and mean opinion score (MOS). 
For MOS evaluation, we use MOSNet for objective evaluation and invite human listeners for subjective evaluation.
Bad case ratio (\%) evaluates the model's robustness by MOS or WER metric.
}
\vspace{0.2cm}
\centering
\resizebox{0.85\columnwidth}{!}{
\begin{tabular}{c|cc|cc|cc}
\toprule[1.2pt]
\multirow{2}{*}{Model}  & WER$\downarrow$ & SIM$\uparrow$ & \multicolumn{2}{c}{MOS $\uparrow$ {by}} & \multicolumn{2}{|c}{Bad Case Ratio $\downarrow$} \\ 
& (\%)& (0,1) & {MOSNet} & {Human} & MOS $\leqslant$ 3 & \%WER > 20 \\ \midrule
VoiceCraft & $35.3$ & $0.79$ & $3.36$ & $3.22$ & $27\%$ & $51\%$ \\ \midrule
RIO-DPO &  $11.3$ & $0.92$ & $4.11_{\textcolor{teal}{+0.75}}$ & - & $5\%$ & $17\%$ \\
RIO-ODPO&  $9.2$ & $0.93$ & $4.15_{\textcolor{teal}{+0.79}}$ & - & $5\%$ & $15\%$ \\
UNO-null  & $6.8$ & $0.93$ & $4.20_{\textcolor{teal}{+0.84}}$ & - & $4\%$ & $11\%$ \\
RIO (ours)  & $\mathbf{3.4}$ & $\mathbf{0.96}$ & $\mathbf{4.40_{\textcolor{teal}{+1.04}}}$ & $\mathbf{4.18_{\textcolor{teal}{+0.96}}}$ & $\mathbf{1\%}$ & $\mathbf{4\%}$ \\ \midrule
Ground-Truth & $1.3$ & - & $4.48$ & $4.54$ & $0\%$ & $0\%$ \\
\bottomrule[1.2pt]
\end{tabular}}
\vspace{0.2cm}
\label{table:main}
\end{table}

\begin{figure}
  % \centering
  \begin{minipage}[h!]{0.45\textwidth}
   \centering
   \includegraphics[scale=0.9]{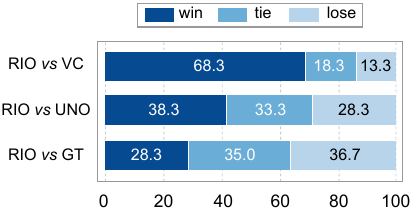}
   \caption{Results of A/B test. ``VC'' and ``GT'' denote the ``VoiceCraft'' and ``Ground-Truth''.}
    \label{f3}
  \end{minipage}
  \hfill
  \begin{minipage}[h!]{0.52\textwidth}
   \centering
   \includegraphics[scale=0.85]{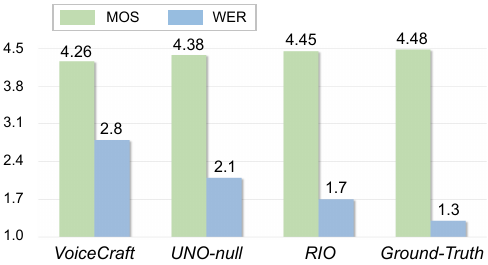}
   % \vspace{-0.1cm}
   \caption{MOS and WER Results on 830M models.}
    \label{f4}
  \end{minipage}
% \vspace{-0.3cm}
\end{figure}

\begin{figure}
  % \centering
    \begin{minipage}[ht]{0.56\textwidth}
    \centering
    \includegraphics[scale=0.255]{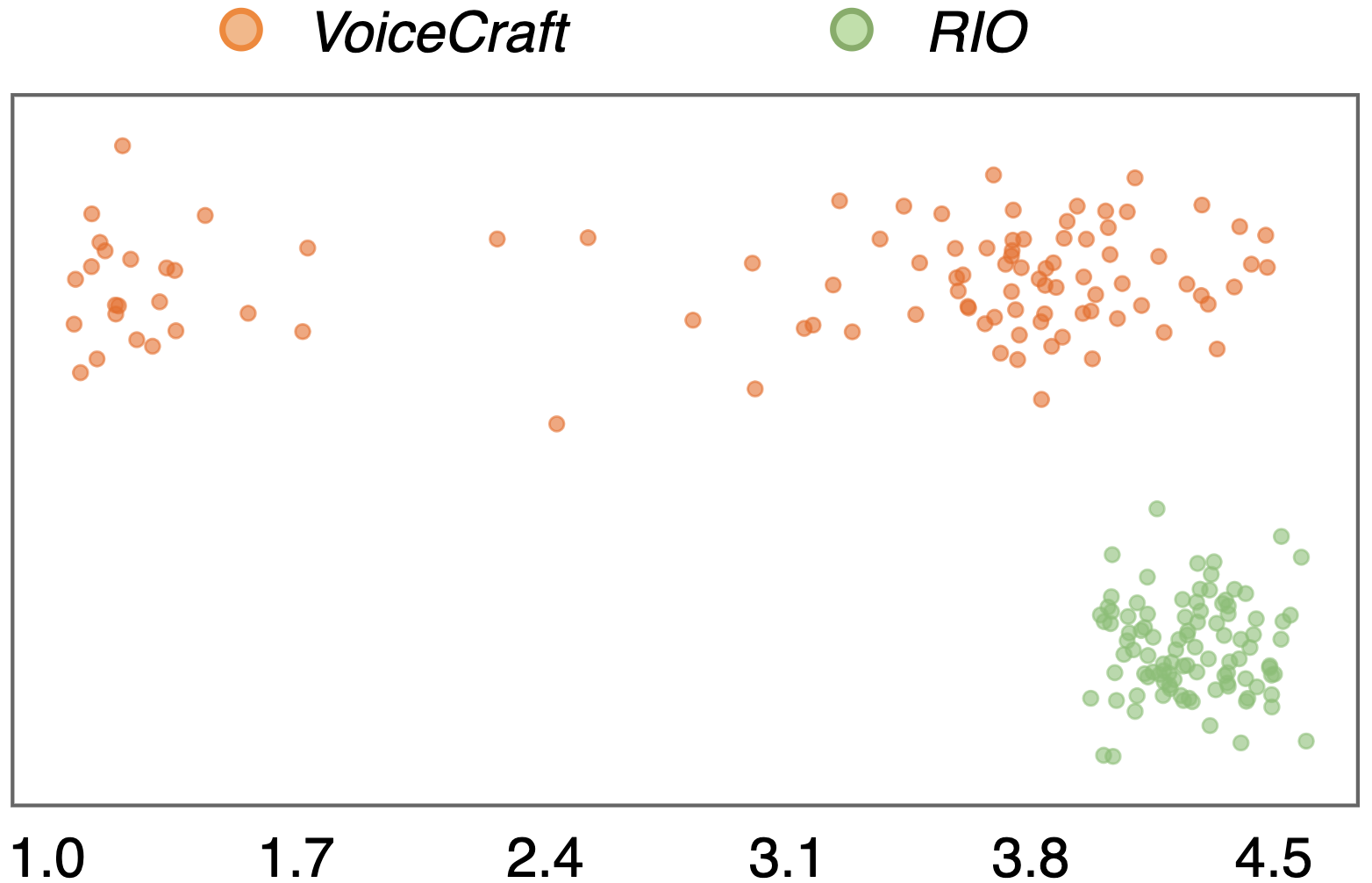}
    \caption{MOS score distributions of \textit{VoiceCraft-330M} baseline and our proposed RIO approach.}
    \label{f5}
    \end{minipage}
  \hfill
    \begin{minipage}[ht]{0.4\textwidth}
    \centering
    \captionof{table}{Ratio of good reverse inference results in all the good zero-shot generations.
    The good TTS results are defined as MOS larger than 3.
    \textit{VoiceCraft} model of different sizes (330M and 830M) are used for statistics.
    ``Baseline'' denotes the original \textit{VoiceCraft} model, and ``RIO'' denotes our optimized model.
    }
    \label{table:mos_vc_rio}
    \scalebox{0.98}{
    \begin{tabular}{c|c|c} \toprule 
    Model & Baseline & RIO  \\ 
    \midrule
    \textit{VoiceCraft-330M} & $54\%$   &   $85\%$ \\ \midrule
    \textit{VoiceCraft-830M}   & $80\%$   &   $97\%$\\
    \bottomrule
    \end{tabular}}
  \end{minipage}
\end{figure}

\subsection{Objective Results}
Table~\ref{table:main} reports the objective results in terms of multiple metrics, including WER, SIM, MOS, and bad case ratio.
Specifically, considering the high cost of human annotation, we employ the pre-trained {MOSNet} to estimate MOS, which has demonstrated excellent capability of cross-domain generalization~\cite{cooper2022generalization}.
From Table~\ref{table:main}, we observe that: 
(1) RIO significantly enhances the TTS performance of \textit{VoiceCraft} baseline in terms of WER, SIM, and the estimated MOS, which even approaches the corresponding metrics of ground-truth speech.
Moreover, RIO reduces the ratio of bad cases\footnote{Here we set two undesirable conditions as bad case by MOS $\leqslant$ 3 and WER > 20 respectively.} to nearly 0\% (1\% by MOS and 4\% by WER on the 330M backbone; 0\% by MOS and 0\% by WER on the 830M backbone), such finding indicates that RIO poses a significant contribution to improving the robustness of zero-shot TTS.
(2) Our reproduced \textit{RIO-DPO} baselines also improves the \textit{VoiceCraft} backbone but yields limited effectiveness.
This limitation stems from the sampling process where the large MOS gap between paired positive and negative samples may not guarantee bad enough quality of the latter.
For example, a good zero-shot generation with a bad reverse inference result could be selected as the negative sample but it is not ``bad'' enough.
However, since DPO requires paired positive and negative samples from the same input, it is hard to generate abundant data with ``sufficiently bad'' negative samples.
(3) The \textit{UNO-null} baseline serves as a special version of RIO without reverse inference in the sampling process. Therefore, our improvement over {UNO-null} indicates the effectiveness of reverse inference in selecting ``robust'' positive samples for optimization.

\subsection{Human Evaluation}
Apart from objective MOS, we also conduct subjective MOS scoring and A/B testing by human listeners to verify the performance gains, and the results are reported in Table~\ref{table:main} and Fig.~\ref{f3}.
Similar to the objective MOS estimations by {MOSNet}, human evaluations also impose a significant gain of our proposed RIO over the vanilla {VoiceCraft}, which well verifies its efficacy. 
Furthermore, the A/B testing results in Fig.~\ref{f3} show that RIO significantly strengthens the robustness of {VoiceCraft} and {UNO-null} baselines, and it even produces comparable speech quality to the ground-truth speech.

\subsection{Scalability to Larger Backbone Models}
To further evaluate the scalability of RIO to larger TTS backbones, we also conduct experiments on {VoiceCraft-830M} and the results are presented in Fig.~\ref{f4}.
We first observe that the 830M version of {VoiceCraft} shows better robustness (\textit{i.e.}, 4.26 MOS and 2.8\% WER) than {VoiceCraft-330M}.
Based on this strong backbone, RIO still produces significant gains and even nearly achieves the upper-bound performance of ground-truth speech (4.45 {vs.} 4.48 in MOS).
Regarding the intelligibility of synthesized speech, RIO also yields a considerable reduction in WER from 2.8\% to 1.7\%, which also approaches the 1.3\% WER of ground-truth speech.
The principle behind this is that RIO can select the ``really robust'' samples from the high-quality generations of {VoiceCraft-830M} via reverse inference, and thus further optimizes the robustness of VoiceCraft.
In summary, our proposed RIO presents remarkable scalability and generality to large backbone models, which well aligns with the current popularity of large speech models in both academia and industry.

\subsection{Analysis of Zero-shot TTS Robustness}
Fig.~\ref{f5} visualizes the MOS distributions of synthesized speech from {VoiceCraft} model with and without RIO optimization.
It can be observed that vanilla {VoiceCraft} generates a considerable number of bad cases and the variance of the MOS of good cases is also large.
This phenomenon indicates the limited robustness of such a backbone model.
In comparison, RIO can remove those bad cases and synthesize high-quality as well as ``robust'' speech samples, which reveals the effectiveness of PPC. 

To further investigate the robustness of zero-shot TTS, reverse inference is performed on the good synthesized speech by {VoiceCraft} and {RIO} to predict their corresponding prompt speech.
Then, the ratio of good reverse inference samples is counted in Table~\ref{table:mos_vc_rio}, which satisfies PPC according to Section~\ref{ssec:reverse_infer}.
It is observed that the 330M baseline presents limited robustness with only 54\% good reverse inference samples, whereas RIO increases this ratio considerably to 85\%.
Moreover, the larger 830M baseline is more robust with 80\% PPC samples that are capable of good reverse inference.
Impressively, RIO can further raise this ratio to nearly 100\%, which shows its remarkable ability to enhance the robustness of zero-shot TTS.

\section{More Discussions and Future Work}
\textbf{Number and Ratio of $\bm{\mathcal{P}_{\text{pos}}}$ and $\bm{\mathcal{P}_{\text{neg}}}$.}
RIO can handle data imbalance in training examples provided by $\mathcal{P}_{\text{pos}}$ and$\mathcal{P}_{\text{neg}}$. In our experiments, we found that both positive and negative samples can provide good optimization effects as long as they meet a certain threshold, which requires approximately only 150 samples, and extra samples only generate marginal improvements. This can be attributed to two factors. First, MOSNet is only used as a binary discriminator in our experiments, which limits its performance in providing rich supervised information. Second, RIO is based on the pre-trained model and its own generated samples, which allows the model to converge rapidly to a local optimum without changing the model parameters significantly.

\textbf{Can TTS Model Understand What It Generated?} The synthesized speech exhibits high sound quality, and in subjective evaluations, some generated samples are even difficult to tell from the ground truth. However, when using the synthesized speech as a speech prompt for second-order decoding, the quality of the newly generated speech $\hat{\mathbf{X}}$ is obviously inferior compared to the original speech prompt $\mathbf{X}$. In this paper, the proposed reverse inference predicts only three seconds of the original speech prompt, while further efforts are required to explore the generation of longer speech segments. This provides a new perspective of PPC for improving TTS quality: a good speech sample synthesized by the TTS should also be able to be perceived by the same TTS model as a prompt for further generating new arbitrary speech samples.
%a good synthesized speech sample should also be comprehensible to the TTS model and capable of prompting new generations.

\textbf{How Can RIO Mitigate the Train-test Mismatch Problem in Auto-regressive Codec TTS?}
As discussed in Section~\ref{sec:intro}, the codec language model usually suffers from the typical train-test mismatch problem in auto-regressive models (\textit{i.e.}, exposure bias~\cite{bengio2015scheduled}).
Consequently, the codec language model is sometimes not certain enough on its prediction, which leads to diverse generations by multinomial sampling and thus increases the chance of bad cases due to error accumulation~\cite{pang2020text,gu2023minillm}.
RIO proposes a sampling-annotating-inference pipeline, which first samples positive and negative exemplars from model generations, and then sends them back into the language model for activation or suppression.
%Therefore, RIO aligns the teacher-forcing training to the auto-regressive decoding, while deviating from those undesirable ones. Specifically, the proposed concept of \textit{reverse inference} further selects more representative exemplars to enhance the robustness of zero-shot TTS. As a result, the codec language model becomes more confident about good predictions and thus generates higher-quality synthesized speech.
Therefore, RIO reduces the mismatch in decoding history between the teacher-forcing training and the test-time auto-regressive decoding, and deviates the model from generating undesirable samples. 
%.while deviating from those undesirable ones. 
Specifically, the proposed concept of \textit{reverse inference} selects more representative exemplars to enhance the robustness of zero-shot TTS.
As a result, the codec language model becomes more confident about those good predictions and thus generates higher-quality synthesized speech.

\textbf{Is RLHF Post-training the Best Solution to Enhance TTS Robustness?}
Though effective, there still exist limitations in enhancing TTS robustness with RLHF post-training.
For example, there exist infinite possibilities of ground-truth speech in the TTS task, but we only optimize the TTS generation towards one single target during both TTS training and RLHF post-training stages, which could limit the robustness of the TTS model.
Therefore, in future work, we hope to find a policy to select universal robust exemplars for RLHF optimization or integrate the RLHF post-training into the initial TTS training stage to implement an end-to-end, universal, and robust TTS optimization scheme.

\textbf{Can We Perform RIO Iteratively to Further Enhance the Robustness?}
There are two directions that we can move on in RIO for iterative optimization.
(1) \textit{``Sampling - RIO - Sampling - RIO - ...''} pipeline: we can employ the RIO optimized model to repeat the sampling and RIO optimization processes. With a robust TTS model, we can sample higher-quality positive exemplars to further refine the TTS model for stronger robustness.
(2) \textit{``Zero-shot Inference - Reverse Inference - Zero-shot Inference - Reverse Inference - ...''}: we can employ multiple-round zero-shot and reverse inference processes to select more robust positive samples for RIO optimization.

\section{Conclusion}
In this work, we present an RLHF-based optimization approach called RIO tailored to the zero-shot TTS system with neural codec modelling. RIO is novel for its reverse inference that aims to select PPC speech samples to enable the use of TTS-generated speech samples as prompt for further TTS generation. 
%to select representative exemplars from self-generated samples, 
%and then use these exemplars to update the TTS model towards the direction of robust zero-shot synthesis.
Updating the TTS model using RLHF with these PPC samples helps to enhance the TTS robustness. 
Experimental results show that without human annotation, RIO effectively enhances the capacity of an advanced zero-shot TTS system and surpasses other optimization baselines in terms of MOS, WER, and SIM, as well as considerably reduces the ratio of bad cases. Moreover, RIO proposes a new insight to analyze the robustness of zero-shot TTS inference, thereby providing a potential scheme for the deployment of the TTS system in practice.

% \clearpage
% \section*{References}
\bibliographystyle{plain}
\bibliography{references}

%%%%%%%%%%%%%%%%%%%%%%%%%%%%%%%%%%%%%%%%%%%%%%%%%%%%%%%%%%%%
\newpage

\end{document}